\documentclass[letterpaper, 10 pt, conference]{ieeeconf}  

\IEEEoverridecommandlockouts                              

\usepackage{cite}
\usepackage{amsmath,amssymb,amsfonts}
\usepackage{algorithmic}
\usepackage{graphicx}
\usepackage{textcomp}
\usepackage{xcolor}
\usepackage{subfig}
\usepackage{tabularx}
\usepackage{booktabs}
\usepackage{multirow}
\usepackage{siunitx}

\makeatletter
\let\NAT@parse\undefined 
\makeatother
\usepackage{hyperref}
\usepackage{eso-pic}
\AddToShipoutPictureBG*{%
  \AtPageLowerLeft{%
    \put(\LenToUnit{0.5\paperwidth},\LenToUnit{0.9cm}){%
      \makebox[0pt][c]{\parbox{0.85\paperwidth}{\centering\scriptsize
        This work has been submitted to the IEEE for possible publication. Copyright may be transferred without notice, after which this version may no longer be accessible.}}}}}

\title{\LARGE \bf
GPU-Accelerated Path-Dependent Marginal Information Gain for Autonomous Exploration}

\author{João Félix Mendes,
Rodrigo Ventura, and
Meysam Basiri
\thanks{This work was supported by Aero.Next project (PRR - C645727867-00000066) and LARSyS FCT funding (DOIs: 10.54499/LA/P/0083/2020, 10.54499/UIDP/50009/2020 and 10.54499/UIDB/50009/2020).}
\thanks{The authors are with the Institute for Systems and Robotics, Department of Electrical and Computer Engineering,
Instituto Superior Técnico, Lisboa, Portugal. {Corresponding Author: João Félix Mendes (e-mail: joao.felix.mendes@tecnico.ulisboa.pt).}
}
}

\begin{document}

\maketitle
\thispagestyle{empty}
\pagestyle{empty}

\begin{abstract}
Autonomous exploration demands that robots continuously evaluate candidate viewpoints based on their expected information gain and execution cost. Sampling-based planners estimate this gain by volumetric raycasting and, due to its computational cost, evaluate candidates under an assumption of mutual independence, ignoring the overlap between viewpoints along the same path. This work presents a GPU-accelerated method for computing path-dependent marginal information gain, where instead of storing and merging the observed unknown voxels along each candidate path, previous observations are represented using depth buffers. Candidate rays are projected into the depth buffers of their ancestors to identify observation overlap and exclude regions expected to be observed. The planning tree is evaluated in depth order to maintain the dependency between viewpoints and their optimized yaws, while candidate nodes and rays at each level are processed in parallel on the GPU. The proposed method stays within $5$-$10\%$ of the exact marginal gain computed using voxel hash maps, with speed-ups of up to $118\times$ on a desktop GPU and $28\times$ on an NVIDIA Jetson Orin NX. The method was integrated into two sampling-based exploration planners and evaluated in three simulation environments, where marginal gain reduced the time to $95\%$ coverage in five of the six evaluated planner-environment combinations. Real-world experiments also showed a $30\%$ reduction in the time to $95\%$ coverage, as well as earlier exploration termination times.
\end{abstract}

\section{INTRODUCTION} \label{sec:introduction}

\begin{figure}[t]
  \centering
  \subfloat[Absolute gain]{%
    \label{subfig:patio_absolute}%
    \includegraphics[width=0.92\columnwidth]{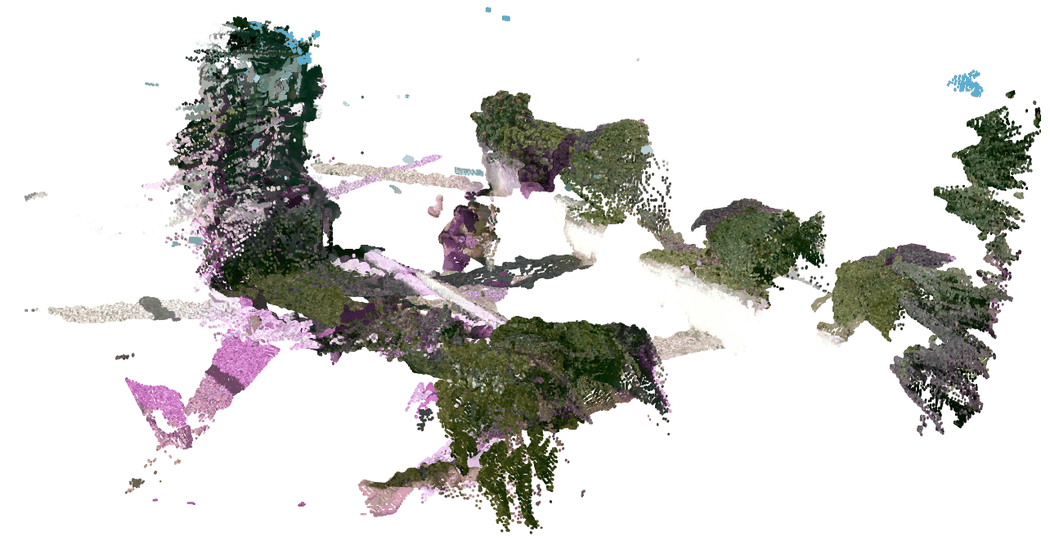}}%
  \\[2pt]
  \subfloat[Marginal gain]{%
    \label{subfig:patio_marginal}%
    \includegraphics[width=0.92\columnwidth]{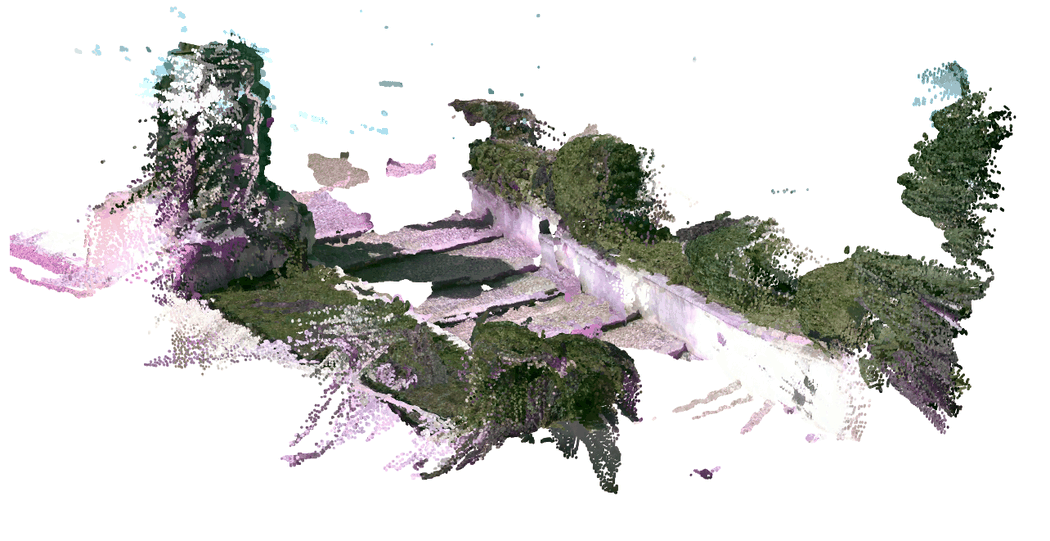}}%
  \caption{Reconstruction of the Patio environment after $150$~s using AEP with absolute gain (a) and marginal gain (b).}
  \label{fig:patio_reconstruction}
\end{figure}

Autonomous exploration of unknown environments is a fundamental problem in robotics, with applications ranging from search-and-rescue operations \cite{Dang_2020} to large-scale environmental exploration \cite{Petracek_2021}. In these scenarios, a robot must continuously select actions that enable efficient mapping of the environment under constraints such as limited sensing, flight time, and onboard computational resources. This decision-making process is referred to as the Next-Best-View problem.

Sampling-based exploration strategies, including the Receding-Horizon Next-Best-View Planner (RH-NBVP) \cite{Bircher_2016}, the Autonomous Exploration Planner (AEP) \cite{Selin_2019}, and related methods \cite{Mendes_2026, Schmid_2020, Lindqvist_2024}, address this problem by iteratively constructing a Rapidly-Exploring Random Tree (RRT) in the robot’s configuration space and evaluating sampled candidates according to their expected information gain. This gain is typically defined as the volume of previously unknown space visible from a candidate viewpoint. The best branch is selected using an objective function that combines information gain and traversal cost (e.g., flight time or path length), and the first segment of that branch is executed.

Information gain is commonly estimated through volumetric raycasting \cite{Woo_1987}, at a computational cost that grows with the number of candidates, sensor resolution, and map resolution. Sampling-based planners evaluate hundreds to thousands of candidates per replanning iteration, so to maintain real-time performance they subsample rays, sample fewer candidates, or coarsen the map \cite{Schmid_2020,Lindqvist_2024}, and evaluate each candidate independently, assuming mutual independence between viewpoints \cite{Bircher_2016,Selin_2019,Mendes_2026,Schmid_2020}. We refer to this per-viewpoint estimate as absolute gain. Since it disregards the observations from preceding viewpoints along the same path, accumulating absolute gain along a branch can count the same unknown regions multiple times, overestimating the information the path provides and changing the ranking of candidate branches.

The information a viewpoint actually contributes is its path-dependent marginal gain, which discounts the space that the preceding viewpoints on the path are already expected to observe. However, computing it exactly is costly. In an exact implementation, the visible unknown voxels of each node are stored in a hash map and, for each candidate, the hash maps of its ancestors are merged before the candidate observation is cross-checked against the merged set. This scales poorly with branch depth, as more ancestor sets must be merged, and with map resolution, as the voxel count grows cubically with decreasing voxel size, making marginal gain difficult to compute in real time over large planning trees.

In this work, we address this limitation with a GPU-accelerated method for path-dependent marginal gain that changes how previous observations are represented. Instead of storing and merging the visible voxels of each ancestor, each node stores a depth buffer, and candidate rays are projected into the buffers of their ancestors to identify and exclude the overlap from the gain. The volumetric set operations become per-ray queries in image space, which parallelize efficiently. The CPU builds the planning tree, performs collision checking, and handles the exploration logic, while the GPU evaluates its information gain. Since each node depends on its ancestors, the tree is processed in depth order, parallelizing same-level nodes and rays.

We evaluate the proposed method in three stages. First, we compare its gain estimates and computation time against the exact marginal gain obtained using voxel hash maps over a range of tree sizes and map resolutions, on a desktop GPU and an NVIDIA Jetson Orin NX. Second, we integrate it into RH-NBVP and AEP and measure exploration performance in three simulated environments, where the effects of tree size and execution horizon are also studied. Finally, we deploy the system onboard an Unmanned Aerial Vehicle (UAV) to validate it under real-world exploration conditions. The framework is open source\footnote{\url{https://github.com/IRSg-ARG/UAV_3d_reconstruction}}. The main contributions are:
\begin{itemize}
    \item A GPU-accelerated method for computing path-dependent marginal gain using depth buffers.
    \item A depth-ordered tree evaluation strategy that preserves the dependency between ancestor observations and yaws while parallelizing candidate nodes and rays within each tree level.
    \item An extensive evaluation of the proposed method, including its accuracy and computational scaling against exact voxel-set marginal gain, its effect on exploration performance across two sampling-based planners, and real-world validation onboard a UAV.
\end{itemize}

\section{RELATED WORK} \label{sec:related_work}

Exploration approaches can be divided into frontier-based, sampling-based, and hybrid planning methods. Frontier-based exploration defines frontiers as the boundary between free and unknown space and guides the robot toward them until the environment is fully mapped \cite{Yamauchi_1997}, whereas hybrid methods combine a local planner that selects and reaches viewpoints with a global planner that guides the overall exploration. FUEL \cite{Zhou_2021} uses frontiers as exploration targets, computes a global visitation order over them by solving an Asymmetric Traveling Salesman Problem, and refines the corresponding viewpoints and trajectories locally. FALCON \cite{Zhang_2025} extends the global stage beyond the detected frontiers by decomposing the environment into connected regions and computing a coverage path over the unexplored space.

Sampling-based approaches follow a different strategy by directly sampling candidate viewpoints or trajectories in free space. The RH-NBVP \cite{Bircher_2016} introduced this approach by iteratively constructing an RRT, evaluating candidate viewpoints using an information-theoretic objective function, and executing the first segment of the highest-scoring branch in a receding-horizon fashion. AEP \cite{Selin_2019} extends this approach with a second-layer global frontier-based planner, activated when in low-information regions, that uses previously sampled informative nodes to escape local minima. In \cite{Schmid_2020} an RRT$^*$ is expanded and maintained during the full mission for global exploration, while other approaches \cite{Lindqvist_2024,Mendes_2026} consider trajectory feasibility and robot kinodynamics during candidate evaluation. We build on this family of planners and address how the sampled candidates are evaluated.

Information gain determines how sampling-based planners rank candidate viewpoints and paths. It is estimated by raycasting through occupancy \cite{Hornung_2013} or Truncated Signed Distance Field (TSDF) \cite{oleynikova2017voxblox} maps and measuring the unknown volume visible from a candidate viewpoint, which is computationally expensive and scales poorly with sensor and map resolution, especially on the CPU. To keep this tractable, existing methods subsample rays and evaluate viewpoints assuming mutual independence between them \cite{Bircher_2016,Selin_2019,Mendes_2026,Schmid_2020}. GPUs have also been used to accelerate volumetric mapping \cite{Millane_2024}, and information gain evaluation in trajectory planning \cite{Renz_2025}. In the latter, candidate gains are evaluated independently, allowing them to be processed in parallel. Marginal gain removes this assumption, making a node's gain depend on the observations and selected yaws of all preceding viewpoints along its path. Therefore, candidates can no longer be evaluated in arbitrary order, reducing the parallelism available for GPU gain evaluation.

Previous work has considered this overlap when evaluating candidate trajectories. ERRT \cite{Lindqvist_2024} computes marginal gain along candidate trajectories, removing unknown voxels already observed at preceding evaluation points. Its implementation relies on UFOMap \cite{Duberg_2020} for efficient volumetric queries and controls the computational cost by limiting the number of candidate branches, evaluating gains only at states separated by a minimum distance $d_{info}$, and by allowing coarser octree depths for gain computation. This makes marginal gain tractable in real time, but only by evaluating a sparser set of states along the trajectory, which omits the information gain at intermediate states. In contrast, we evaluate marginal gain by changing how previous observations are represented and exploiting GPU parallelism. This avoids building a volumetric union of ancestor observations per node while keeping the dependency between viewpoints along each path.

\section{PROBLEM FORMULATION} \label{sec:problem_formulation}

\subsection{Exploration as Sequential Next-Best-View Selection}

Consider an initially unknown static three-dimensional environment $\mathcal{W} \subset \mathbb{R}^3$ and a UAV equipped with a depth camera. The goal of an exploration planner is to generate collision-free feasible paths that build a complete map of $\mathcal{W}$ from the measurements acquired by the perception sensor. Since the environment is not known a priori, exploration is performed online in a receding-horizon fashion, replanning as new observations become available.

At each planning iteration, the current map $\mathcal{W}_{\mathrm{map}}$ partitions the environment into voxels of edge length $v_{\mathrm{size}}$, each labelled as free, occupied, or unknown. A sampling-based planner uses $\mathcal{W}_{\mathrm{map}}$ to construct a tree $\mathcal{T}=(\mathcal{N},\mathcal{E})$ rooted at the UAV's current state, where each edge $e\in\mathcal{E}$ is a collision-free segment connecting two nodes. In turn, each node $n\in\mathcal{N}$ is characterized by a candidate viewpoint position $\mathbf{p}_n\in\mathbb{R}^3$ and yaw $\psi_n$, information gain $g(n)$, cumulative traversal cost $c(n)$, and path objective $o(n)$. The planner ranks the candidate paths according to $o(n)$, executes the first segment of the highest-valued path, and replans.

\subsection{Information Gain Formulations}

For a candidate position $\mathbf{p}_n$ and yaw $\psi$, let $\mathcal{I}(\mathbf{p}_n,\psi)$ be the set of unknown voxels visible from that viewpoint. Existing approaches \cite{Bircher_2016,Selin_2019,Schmid_2020,Mendes_2026} evaluate candidate viewpoints independently using absolute gain, defined as
\begin{equation}
    g_{\mathrm{abs}}(\mathbf{p}_n,\psi)
    =
    v_{\mathrm{size}}^{3}
    \left|\mathcal{I}(\mathbf{p}_n,\psi)\right|,
    \label{eq:absolute_gain_yaw}
\end{equation}
where $v_{\mathrm{size}}^{3}$ is the voxel volume and $\left|\mathcal{I}(\mathbf{p}_n,\psi)\right|$ denotes the number of visible unknown voxels.

The yaw $\psi_n$ is selected by maximizing $g_{\mathrm{abs}}(\mathbf{p}_n,\psi)$ over $\psi\in\Psi$, and the resulting gain is stored as $g_{\mathrm{abs}}(n)$. Since this gain depends only on the current viewpoint, overlap with preceding viewpoints along the same path is ignored. The same unknown voxels may contribute to multiple nodes, overestimating the accumulated path gain. As overlap differs between paths, this can also change their relative ranking.

To eliminate the mutual independence assumption, the gain of a node must be conditioned on the observations expected from the previous viewpoints along its path. Let $\mathrm{Pa}(n)$ be the parent of node $n$ and $\mathcal{A}(n)=\{\mathrm{Pa}(n), \mathrm{Pa}(\mathrm{Pa}(n)),\ldots,n_{\mathrm{root}}\}$ its ancestor set, containing all preceding nodes along the path to $n$. Marginal gain accounts for all ancestors,
\begin{equation}
    g_{\mathrm{all}}(\mathbf{p}_n,\psi)
    =
    v_{\mathrm{size}}^{3}
    \left|
        \mathcal{I}(\mathbf{p}_n,\psi)
        \setminus
        \bigcup_{a\in\mathcal{A}(n)}
        \mathcal{I}(\mathbf{p}_a,\psi_a)
    \right|.
    \label{eq:path_dependent_gain}
\end{equation}
Similar to the absolute gain, the yaw $\psi_n$ is selected by maximizing $g_{\mathrm{all}}(\mathbf{p}_n,\psi)$ over $\psi\in\Psi$, and the resulting gain is stored as $g_{\mathrm{all}}(n)$. Both the gain and yaw depend on the ancestor observations and selected yaws. The nodes must therefore be evaluated in tree-depth order. The two formulations satisfy $g_{\mathrm{all}}(n)\leq g_{\mathrm{abs}}(n)$.

\subsection{Cost and Objective Formulation}

Candidate paths are ranked by an objective function that combines information gain and traversal cost. The cumulative cost $c(n)$ to node $n$ is
\begin{equation}
    c(n)
    =
    c(\mathrm{Pa}(n))
    +
    \left\|
        \mathbf{p}_n-\mathbf{p}_{\mathrm{Pa}(n)}
    \right\|,
    \label{eq:traversal_cost}
\end{equation}
corresponding to the traveled distance along the path. We use the exponentially decaying objective commonly adopted in sampling-based planners \cite{Bircher_2016,Selin_2019,Mendes_2026},
\begin{equation}
    o_{\mathrm{exp}}(n)
    =
    o_{\mathrm{exp}}(\mathrm{Pa}(n))
    +
    g(n)e^{-\lambda c(n)},
    \label{eq:exponential_objective}
\end{equation}
where $\lambda>0$ is a penalizing coefficient for traversal cost. This favors informative paths while reducing the contribution of viewpoints that are farther from the tree root.

\section{PROPOSED APPROACH} \label{sec:proposed_approach}

\begin{figure} [t]
  \centering
  \includegraphics[width=0.95\columnwidth]{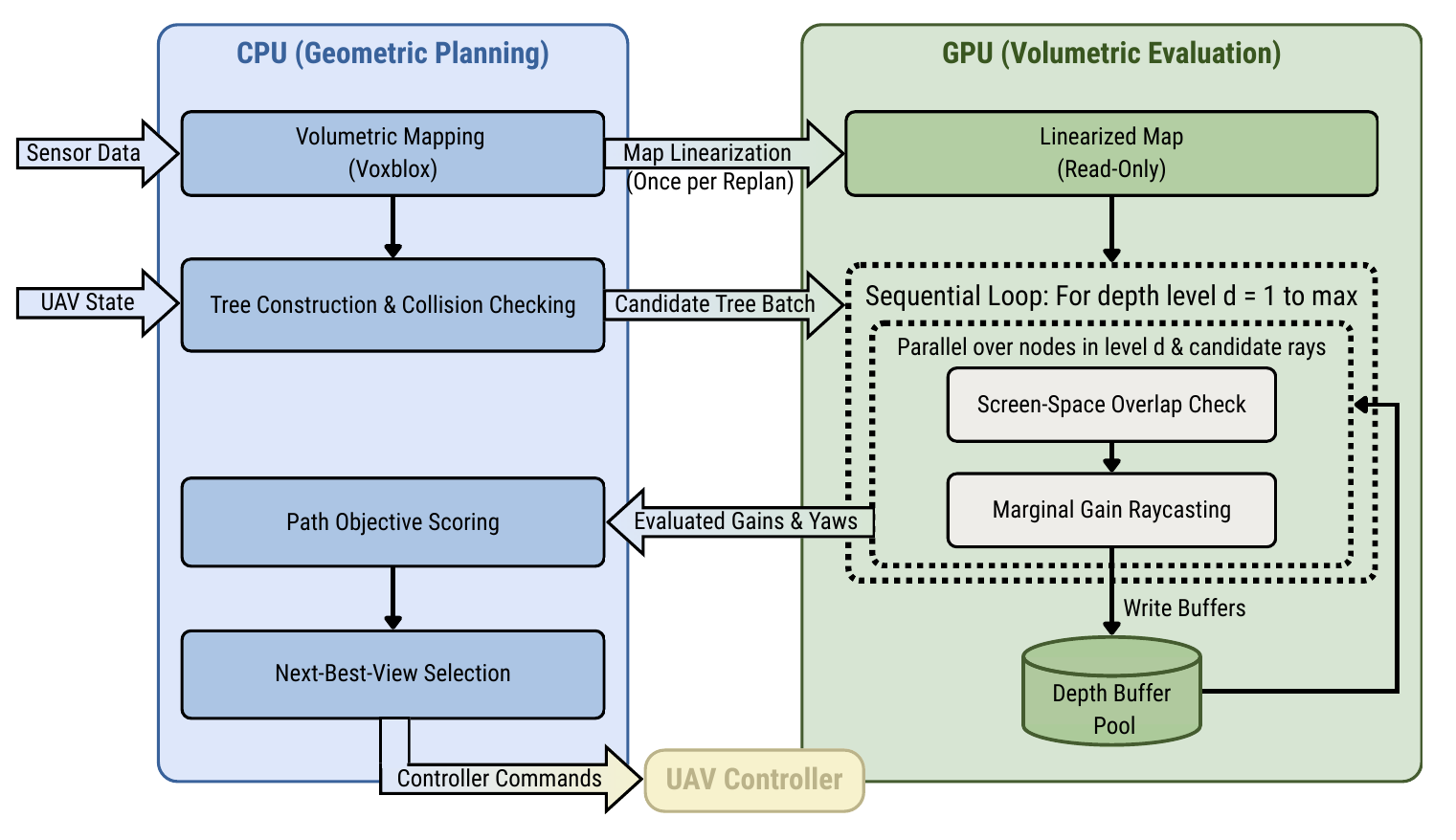}
  \caption{System architecture of the proposed framework. The CPU manages geometric planning and mapping, while the GPU evaluates the candidate tree in depth order, parallelizing nodes and rays within each depth level and storing the resulting depth buffers for descendant evaluation.}
  \label{fig:system_overview}
\end{figure}

We propose a GPU-accelerated method for computing path-dependent marginal gain in sampling-based exploration, built on the hybrid CPU-GPU framework shown in Fig.~\ref{fig:system_overview}. The CPU handles geometric planning, including tree construction, collision checking, path scoring, branch selection, and replanning, while the GPU evaluates the information gain. For this, the CPU-based volumetric map \cite{oleynikova2017voxblox} is first converted into a contiguous representation and transferred to the GPU. The GPU stores ancestor observations as depth buffers, evaluates their overlap with candidate observations in screen space, optimizes the candidate yaw, and computes the gain. The tree is processed sequentially by depth level, with nodes and rays within each level evaluated in parallel. The resulting gains and yaws are returned to the CPU for path scoring and branch selection. 

\subsection{Depth Buffer Representation of Viewpoint Observations}

Each node $n$ stores its viewpoint observation as a depth buffer $D_n\in\mathbb{R}^{W_D\times H_D}$, replacing the voxel hash map used by an exact implementation. Each pixel stores the camera-frame $z$-depth of the first occupied voxel along its ray, or of the ray endpoint at the maximum sensing range $d_{\max}$ if no occupied voxel is hit. Free and unknown voxels are traversed without terminating the ray.

The buffer resolution is chosen so that one pixel at the maximum sensing range matches the voxel size $v_{\mathrm{size}}$. For horizontal and vertical fields of view $\theta_h$ and $\theta_v$, we set $W_D= \left \lceil 2d_{\max}\tan(\theta_h/2)/v_{\mathrm{size}} \right \rceil$ and $H_D= \left\lceil 2d_{\max}\tan(\theta_v/2)/v_{\mathrm{size}} \right\rceil$. Once $\psi_n$ is selected, its depth buffer is kept on the GPU and used to evaluate descendants.

\subsection{Marginal Information Gain Evaluation}

\begin{figure}[t]
  \centering
  \includegraphics[width=0.95\columnwidth]{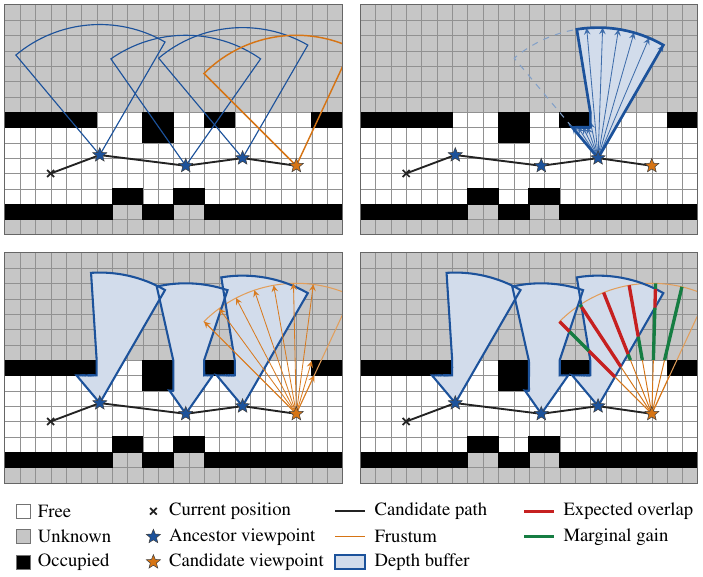}
  \caption{Marginal gain calculation using depth buffers in 2D. \textbf{Top left:} candidate branch of the planning tree shown by the black path, with the candidate viewpoint in orange, its ancestors in blue, and their fields of view. \textbf{Top right:} depth buffer of an ancestor viewpoint, where each ray stores the camera-frame $z$-depth of the first occupied voxel. \textbf{Bottom left:} depth buffers of the ancestor viewpoints and the projected raycast of the candidate viewpoint. \textbf{Bottom right:} projection of the candidate rays into the ancestor depth buffers, identifying observed segments (red) and newly visible segments (green).}
  \label{fig:marginal_gain_2d}
\end{figure}

Information gain is evaluated by raycasting from $(\mathbf{p}_n,\psi)$ in the current map. For marginal gain, the portions of each ray already observed by the ancestors of $n$ are identified using their depth buffers. To determine this overlap, the candidate ray is projected into each ancestor buffer and its depth is compared with the stored values. The camera is modeled using a pinhole camera model to perform this comparison in screen space. 

Consider a candidate ray $\mathbf{r}(t)=\mathbf{p}_n+t\mathbf{d}$, with $0\leq t\leq d_{\max}$, where $\mathbf{d}$ is its direction. For each ancestor $a\in\mathcal{A}(n)$, the ray origin and direction are transformed into the ancestor camera frame and the ray is clipped along the camera $z$-axis to the valid depth range. The resulting 3D segment is projected onto the ancestor depth buffer as
\begin{equation}
    u=f_x\frac{x}{z}+c_x,
    \qquad
    v=f_y\frac{y}{z}+c_y,
    \label{eq:pinhole_projection}
\end{equation}
where $f_x$ and $f_y$ are the focal lengths and $(c_x,c_y)$ is the principal point. The resulting 2D segment is clipped to the depth buffer bounds using the Liang-Barsky algorithm \cite{Liang_1984}. Together, the depth and image clipping restrict the overlap check to the valid camera frustum and, with the buffer resolution defined previously, keep the spacing between samples of the projected ray smaller than or equal to voxel size. Without this clipping, segments outside the sensing range or field of view would be undersampled in screen space, introducing errors in the depth interpolation and overlap estimation.

The clipped segment is traversed using a 2D digital differential analyzer (DDA) \cite{Woo_1987}. Under perspective projection, depth is nonlinear in screen space, but inverse depth $w=1/z$ is linear. So, for interpolation parameter $s\in[0,1]$,
\begin{equation}
    w(s)=(1-s)w_0+sw_1,
    \qquad
    z(s)=\frac{1}{w(s)},
    \label{eq:inverse_depth}
\end{equation}
where $w_0$=$1/z_0$ and $w_1$=$1/z_1$ are the inverse depths at segment endpoints. At each visited pixel, $z(s)$ is compared with the depth in $D_a$, and the segment is considered previously observed when $z(s)$ is smaller or equal to the stored depth.

Since this condition can change several times along the segment, one ancestor can produce multiple overlap intervals,
\begin{equation}
    \mathcal{S}_a(\mathbf{r})
    =
    \left\{
        [t_{a,1}^{\mathrm{in}},t_{a,1}^{\mathrm{out}}],
        \ldots,
        [t_{a,K_a}^{\mathrm{in}},t_{a,K_a}^{\mathrm{out}}]
    \right\},
    \label{eq:ancestor_intervals}
\end{equation}
where $K_a$ is the number of overlap intervals from ancestor $a$. The ancestors are processed sequentially, merging each new $\mathcal{S}_a(\mathbf{r})$ with the accumulated intervals when they overlap.

Once the overlap intervals are determined, a 3D DDA computes the gain. Voxels inside the merged intervals are traversed but do not contribute, while unknown voxels outside them contribute normally and occupied voxels terminate the ray traversal. Figure~\ref{fig:marginal_gain_2d} illustrates this process.

\subsection{Depth-Level Parallel Evaluation}

Marginal gain introduces an ordering constraint because the gain and selected yaw of a node depend on its ancestors. However, this constraint is limited to tree depth, since nodes at the same depth cannot be ancestors of one another. Consequently, once all shallower levels are processed, the nodes at the current depth can be evaluated independently. 

Let $\mathcal{L}_d=\{n\in\mathcal{N}:\operatorname{depth}(n)=d\}$ be the set of nodes at depth $d$. The GPU evaluates the tree in increasing depth order. For each $n\in\mathcal{L}_d$, all candidate yaws are evaluated against the depth buffers of its ancestors and the yaw maximizing the gain is selected. Nodes and sensor rays within the same level are evaluated in parallel and, after each level, the resulting gains and yaws are returned to the CPU.

Therefore, the evaluation is sequential across depth levels but parallel within each level, preserving the ancestor ordering of marginal gain while still exploiting GPU parallelism.

\subsection{Integration with Exploration Planners}

Conventional sampling-based planners evaluate each node as it is generated. However, this is not suitable for a batched, depth-ordered evaluation like the one proposed. Therefore, in our framework, the CPU first builds the complete candidate tree by sampling and collision checking, without evaluating gain during tree expansion. The tree is then organized by depth and evaluated on the GPU, which returns the gain and selected yaw of each node to the CPU. Finally, the CPU computes the traversal costs and path objectives and selects the best branch to execute. By delaying gain evaluation until the tree is complete, enough candidates are provided to use the GPU parallelism. This structure is integrated into RH-NBVP \cite{Bircher_2016} and AEP \cite{Selin_2019}, whose sampling, collision checking, path scoring, branch selection and replanning logic are unchanged. Only the per-node gain evaluation is replaced by the depth-ordered GPU evaluation of the complete tree.

\subsection{Computational Complexity}

The 3D DDA used to compute information gain is common to both the exact voxel hash map implementation and the proposed depth buffer method. As such, we only compare the additional time and space complexity needed to condition a candidate observation on its ancestors.
Let $A_n=|\mathcal{A}(n)|$ be the number of ancestors of node $n$. In the hash map implementation, for a fixed observed volume, the number of unknown voxels in an observation scales as $\mathcal{O}(v_{\mathrm{size}}^{-3})$. Evaluating a candidate requires merging the $A_n$ ancestor maps and checking its voxels against the merged set. Assuming $\mathcal{O}(1)$ average insertion and lookup, the time complexity is $T_{\mathrm{hash}}=\mathcal{O}(A_n v_{\mathrm{size}}^{-3})$. Each stored observation requires $\mathcal{O}(v_{\mathrm{size}}^{-3})$ space, while the merged ancestor map can grow to $\mathcal{O}(A_n v_{\mathrm{size}}^{-3})$. To avoid storing a merged map per node, only local observations are stored, and the merged map is rebuilt when needed.

The proposed method stores one depth buffer per node. With $N_r=W_DH_D$ rays and $W_D,H_D=\mathcal{O}(v_{\mathrm{size}}^{-1})$, each observation requires $N_r=\mathcal{O}(v_{\mathrm{size}}^{-2})$ depth values. Each candidate ray is checked against the $A_n$ ancestor buffers and traverses at most $L_D=\mathcal{O}(W_D+H_D)=\mathcal{O}(v_{\mathrm{size}}^{-1})$ pixels in each buffer. The time complexity is $T_{\mathrm{depth}}=\mathcal{O}(A_n N_r L_D)=\mathcal{O}(A_n v_{\mathrm{size}}^{-3})$.

Both approaches have the same worst-case time complexity, but differ by a factor of $\mathcal{O}(v_{\mathrm{size}}^{-1})$ in space complexity, with the hash map additionally constructing merged ancestor maps during evaluation. A GPU implementation of the hash map approach is possible, but the path-dependent formulation requires a merged set for each candidate with up to $\mathcal{O}(A_n v_{\mathrm{size}}^{-3})$ entries, whose size is unknown in advance and which must be written by multiple threads at once. Our method avoids this, since its $\mathcal{O}(v_{\mathrm{size}}^{-2})$ candidate rays are independent and can be evaluated in parallel, while the $\mathcal{O}(A_n v_{\mathrm{size}}^{-1})$ traversal stays local to each ray.

\section{EXPERIMENTAL EVALUATION} \label{sec:experimental_evaluation}

\begin{table}[t]
\setlength{\tabcolsep}{4pt}
\renewcommand{\arraystretch}{0.85}
\caption{Main simulation parameters.}
\label{tab:simulation_parameters}
\footnotesize
\begin{center}
\begin{tabularx}{\linewidth}{l
    >{\centering\arraybackslash}X
    >{\centering\arraybackslash}X
    >{\centering\arraybackslash}X}
\toprule
\textbf{Parameter} &
\textbf{School} &
\textbf{Large maze} &
\textbf{Multi-story} \\
\midrule
Size [\SI{}{\meter}]
    & $50{\times}35{\times}20$
    & $50{\times}50{\times}3$
    & $24{\times}22{\times}6$ \\
$v_{\mathrm{size}}$ [\SI{}{\meter}]
    & 0.2 & 0.1 & 0.1 \\
Step size [\SI{}{\meter}]
    & 2.0 & 1.5 & 1.0 \\
$N_{\max}$
    & 250 & 250 & 400 \\
$N_{\mathrm{termination}}$
    & 500 & 1000 & 1600 \\
$d_{\mathrm{collision}}$ [\SI{}{\meter}]
    & 1.5 & 0.8 & 0.5 \\
$g_{\mathrm{zero}}$ [\SI{}{\meter\cubed}]
    & 5.0 & 5.0 & 2.0 \\
\bottomrule
\end{tabularx}
\end{center}
\end{table}

The proposed approach is evaluated in simulation using the MRS UAV System \cite{Baca_2021} and Gazebo within ROS Noetic. A simulated UAV uses an Intel RealSense D435i depth camera for perception and Voxblox \cite{oleynikova2017voxblox} for mapping, whose ESDF is used for collision checking, with candidate nodes required to keep a minimum safety distance $d_{\mathrm{collision}}$ from occupied space. Gain evaluation uses a sensor range of \SI{5}{\meter}, a field of view of $87^\circ\times58^\circ$, and a camera pitch of $10^\circ$. Experiments run on a desktop computer with an Intel Core i7-13700K CPU and an NVIDIA GeForce RTX 5060. Embedded performance is evaluated on the NVIDIA Jetson Orin NX with 16~GB of unified memory, the onboard computer used in the real-world experiments.

We evaluate the proposed marginal gain against the absolute gain in RH-NBVP \cite{Bircher_2016} and AEP \cite{Selin_2019}. Each comparison keeps the planner parameters and initial conditions fixed, differing only in the gain formulation. Absolute gain is also evaluated on the GPU so that both formulations use GPU-accelerated gain evaluation. In AEP, we use the global planner score function of \cite{Mendes_2026}, and in RH-NBVP yaw is optimized instead of sampled randomly \cite{Witting_2018}, avoiding random viewpoint overlap between runs. Each condition is evaluated over 10 runs, reported as mean $\pm$ standard deviation. Table~\ref{tab:simulation_parameters} summarizes main parameters used in each environment, where the step size is the maximum edge length, $N_{\max}$ the target tree size, $N_{\mathrm{termination}}$ the maximum number of nodes generated when no informative branch is found, and $g_{\mathrm{zero}}$ the minimum gain for AEP to consider a node informative. Across all experiments $\lambda=0.5$, with $v_{xy_{\max}}=v_{z_{\max}}=\SI{1}{\meter\per\second}$ and $a_{xy_{\max}}=a_{z_{\max}}=\SI{1}{\meter\per\second\squared}$.

Experiments are performed in the school, large maze, and multi-story scenarios. Performance is measured by the time to reach $25\%$, $50\%$, $75\%$, and $95\%$ coverage ($E_{25}$-$E_{95}$), final coverage $C_f$, path length $L$, and average velocity $\bar{v}$. Since AEP self-terminates, its termination time $T_{\mathrm{term}}$ is reported, whereas RH-NBVP runs for a fixed mission duration.

\subsection{Gain Evaluation Accuracy and Computation Time}
\label{sec:gain_evaluation}

\begin{figure}[t]
  \centering
  \subfloat[Information gain]{%
    \label{subfig:gain_depth}%
    \includegraphics[width=0.49\columnwidth]{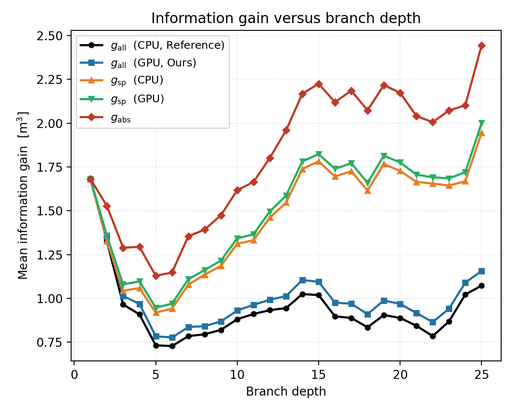}}%
  \hfill
  \subfloat[Relative overestimation]{%
    \label{subfig:gain_overestimate}%
    \includegraphics[width=0.49\columnwidth]{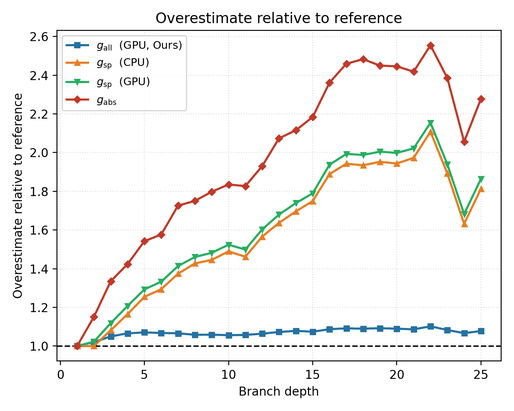}}%
  \caption{Mean information gain as a function of branch depth (a), and its ratio to the exact marginal gain (b). Curves show $g_{\mathrm{all}}$ from the exact CPU hash map implementation and the proposed GPU depth buffer method, with $g_{\mathrm{sp}}$ and $g_{\mathrm{abs}}$.}
  \label{fig:gain_depth}
\end{figure}

\begin{table*}[t]
\centering
\renewcommand{\arraystretch}{0.85}
\caption{Gain evaluation time per replanning iteration [ms], averaged over 10 iterations. Speed-up is the CPU/GPU marginal gain ratio. The resolution factor is the mean increase in evaluation time when reducing $v_{\mathrm{size}}$ from \SI{0.2}{\meter} to \SI{0.1}{\meter}.}
\label{tab:gain_timing}

\footnotesize
\begin{tabular}{cc
                cc cc c
                cc cc c}
\toprule

& &
\multicolumn{5}{c}{\textbf{Desktop}} &
\multicolumn{5}{c}{\textbf{Jetson Orin NX}} \\

\cmidrule(lr){3-7}
\cmidrule(lr){8-12}

& &
\multicolumn{2}{c}{\textbf{Absolute}} &
\multicolumn{3}{c}{\textbf{Marginal}}
&
\multicolumn{2}{c}{\textbf{Absolute}} &
\multicolumn{3}{c}{\textbf{Marginal}}
\\

\cmidrule(lr){3-4}
\cmidrule(lr){5-7}
\cmidrule(lr){8-9}
\cmidrule(lr){10-12}

$\boldsymbol{v_{\mathrm{size}}}$ &
$\boldsymbol{N}$ &
\textbf{CPU} &
\textbf{GPU} &
\textbf{CPU} &
\textbf{GPU} &
\textbf{Speed-up} &
\textbf{CPU} &
\textbf{GPU} &
\textbf{CPU} &
\textbf{GPU} &
\textbf{Speed-up} \\

\midrule

\multirow{6}{*}{\SI{0.2}{\meter}}
& 50
& 40.6 & \textbf{0.8}
& 57.7 & \textbf{9.2}
& \textbf{$\times6$}
& 65.0 & \textbf{3.1}
& 183.6 & \textbf{53.1}
& \textbf{$\times3$} \\

& 100
& 88.6 & \textbf{0.6}
& 201.9 & \textbf{26.9}
& \textbf{$\times8$}
& 112.1 & \textbf{5.4}
& 189.4 & \textbf{61.5}
& \textbf{$\times3$} \\

& 500
& 491.2 & \textbf{1.9}
& 1097.4 & \textbf{46.5}
& \textbf{$\times24$}
& 570.5 & \textbf{24.4}
& 866.5 & \textbf{138.5}
& \textbf{$\times6$} \\

& 1000
& 921.7 & \textbf{3.6}
& 2945.5 & \textbf{70.6}
& \textbf{$\times42$}
& 1197.0 & \textbf{49.2}
& 3145.6 & \textbf{259.9}
& \textbf{$\times12$} \\

& 5000
& 4419.0 & \textbf{15.0}
& 10904.5 & \textbf{225.9}
& \textbf{$\times48$}
& 5459.6 & \textbf{242.5}
& 7879.1 & \textbf{1344.4}
& \textbf{$\times6$} \\

& 10000
& 8260.7 & \textbf{27.7}
& 16297.3 & \textbf{401.8}
& \textbf{$\times41$}
& 10844.2 & \textbf{489.4}
& 25945.1 & \textbf{2547.0}
& \textbf{$\times10$} \\

\midrule

\multirow{6}{*}{\SI{0.1}{\meter}}
& 50
& 270.6 & \textbf{1.9}
& 765.8 & \textbf{62.1}
& \textbf{$\times12$}
& 411.5 & \textbf{12.6}
& 1430.6 & \textbf{142.2}
& \textbf{$\times10$} \\

& 100
& 622.7 & \textbf{3.7}
& 3630.9 & \textbf{97.6}
& \textbf{$\times37$}
& 836.5 & \textbf{26.2}
& 6266.0 & \textbf{379.0}
& \textbf{$\times17$} \\

& 500
& 3367.1 & \textbf{12.2}
& 24783.9 & \textbf{359.7}
& \textbf{$\times69$}
& 4039.2 & \textbf{119.8}
& 17256.2 & \textbf{621.7}
& \textbf{$\times28$} \\

& 1000
& 6594.9 & \textbf{23.9}
& 39086.2 & \textbf{330.7}
& \textbf{$\times118$}
& 7406.3 & \textbf{229.6}
& 17799.9 & \textbf{1214.3}
& \textbf{$\times15$} \\

& 5000
& 29183.1 & \textbf{105.3}
& 118996.1 & \textbf{1020.9}
& \textbf{$\times117$}
& 35340.8 & \textbf{1146.7}
& 84942.9 & \textbf{5787.2}
& \textbf{$\times15$} \\

& 10000
& 55061.0 & \textbf{191.8}
& 181567.4 & \textbf{1950.0}
& \textbf{$\times93$}
& 66733.5 & \textbf{2272.5}
& 123302.6 & \textbf{11925.8}
& \textbf{$\times10$} \\

\midrule

\multicolumn{2}{l}{\textbf{Resolution factor}}
& $6.8\times$
& \textbf{$5.7\times$}
& $14.3\times$
& \textbf{$5.2\times$}
& --
& $6.6\times$
& \textbf{$4.6\times$}
& $10.7\times$
& \textbf{$4.4\times$}
& -- \\

\bottomrule
\end{tabular}
\end{table*}

We first validate the proposed GPU depth buffer marginal gain against the exact CPU hash map implementation using fixed yaw angles so that both evaluate the same viewpoints. Across 166,440 node evaluations, absolute gain matches the reference exactly, while the proposed marginal gain achieves $R^2=0.9937$ with a regression slope of $1.028$. The resulting difference comes because each depth buffer pixel represents the area covered by a beam instead of a ray, and can correspond to more than one voxel, up to four at the resolutions used. When the depth is interpolated, this approximation error accumulates along the beam, resulting in a small overestimation of the gain. Figure~\ref{fig:gain_depth} shows that the proposed method remains close to the exact marginal gain, with a $5$-$10\%$ overestimation. In contrast, single-parent $g_{sp}$ and absolute gain errors increase with branch depth as more ancestor observations accumulate, reaching up to $2\times$ and $2.6\times$ the exact marginal gain, respectively.

Table~\ref{tab:gain_timing} compares the CPU and GPU evaluation times of absolute and marginal gain for different tree sizes $N$ on the desktop and Jetson Orin NX. GPU timings include gain computation and CPU-GPU data transfers and are averaged over 10 replanning iterations. The Jetson timings were obtained using hardware-in-the-loop, with Gazebo running on the desktop while mapping, planning, and gain evaluation ran on the Jetson. On the desktop, the proposed marginal gain is $6$-$48\times$ faster than the hash map implementation at \SI{0.2}{\meter} and $12$-$118\times$ faster at \SI{0.1}{\meter}. For $N=10000$, evaluation time drops from \SI{16.30}{\second} to \SI{0.40}{\second} and from \SI{181.57}{\second} to \SI{1.95}{\second}, respectively. The Jetson shows the same trend, achieving speed-ups of $3$-$12\times$ and $10$-$28\times$. The method also scales better with map resolution since halving the voxel size increases the geometric mean marginal-gain time by $14.3\times$ on the desktop CPU versus $5.2\times$ on the GPU, and by $10.7\times$ versus $4.4\times$ on the Jetson. GPU absolute gain remains faster than marginal gain in every configuration, so any exploration improvement reported in the following sections is obtained despite a higher gain evaluation cost.

\subsection{Effect of Tree Size} \label{sec:tree_size}

We compare AEP and RH-NBVP in the school environment, varying $N_{\max}$ from 50 to 500, with $N_{\mathrm{termination}}$ scaled accordingly. Table~\ref{tab:tree_size} summarizes the results. For both planners, marginal gain improves exploration efficiency, reducing $E_{95}$. With RH-NBVP, the effect becomes more significant as the tree grows. At $N_{\max}=50$ the difference in $E_{95}$ is small, since the limited number of branches often leads both gains to select the same direction. Larger trees provide more alternative branches, making overlap more likely to affect their ranking and change the selected branch. AEP benefits from marginal gain even with the smallest tree. Removing overlapping regions makes low-information branches fall below $g_{\mathrm{zero}}$ earlier, allowing AEP to switch to its global planner sooner. It also improves the decisions of the global planner, which, unlike RH-NBVP that only executes the first segment of the selected branch, executes the full path to the selected waypoint. Therefore, accounting for overlap along that path improves the waypoint selection. This is also reflected in the shorter paths and earlier termination times obtained across all tree sizes.

\begin{table}[t]
\centering
\caption{Effect of tree size in the school environment. A/M denote absolute and marginal gain.}
\label{tab:tree_size}
\footnotesize
\setlength{\tabcolsep}{2.1pt}
\renewcommand{\arraystretch}{0.90}
\begin{tabular}{@{}lllccc@{}}
\toprule
&
&
&
\multicolumn{3}{c}{$\boldsymbol{N_{\max}}$} \\
\cmidrule(lr){4-6}

\textbf{Planner} &
\textbf{Metric} &
\textbf{Gain} &
\textbf{50} &
\textbf{250} &
\textbf{500} \\
\midrule

\multirow{6}{*}{\shortstack{RH-\\NBVP}}
& \multirow{2}{*}{$E_{95}$ [min]}
& A & $20.6{\pm}3.5$ & $21.6{\pm}1.9$ & $23.8{\pm}1.9$ \\
&
& M & $\mathbf{20.3{\pm}2.4}$ & $\mathbf{20.9{\pm}2.7}$ & $\mathbf{21.6{\pm}2.5}$ \\

\cmidrule(lr){2-6}

& \multirow{2}{*}{$L$ [m]}
& A & $1010{\pm}3$ & $\mathbf{913{\pm}6}$ & $\mathbf{806{\pm}20}$ \\
&
& M & $\mathbf{1007{\pm}5}$ & $935{\pm}8$ & $877{\pm}7$ \\

\cmidrule(lr){2-6}

& \multirow{2}{*}{$\bar{v}$ [m/s]}
& A & $0.56{\pm}0.01$ & $0.50{\pm}0.01$ & $0.45{\pm}0.01$ \\
&
& M & $0.56{\pm}0.01$ & $\mathbf{0.52{\pm}0.01}$ & $\mathbf{0.49{\pm}0.01}$ \\

\midrule

\multirow{8}{*}{AEP}
& \multirow{2}{*}{$E_{95}$ [min]}
& A & $16.8{\pm}0.6$ & $18.4{\pm}0.8$ & $21.1{\pm}0.9$ \\
&
& M & $\mathbf{15.0{\pm}0.6}$ & $\mathbf{16.6{\pm}1.1}$ & $\mathbf{18.4{\pm}1.5}$ \\

\cmidrule(lr){2-6}

& \multirow{2}{*}{$L$ [m]}
& A & $908{\pm}88$ & $895{\pm}29$ & $829{\pm}39$ \\
&
& M & $\mathbf{808{\pm}36}$ & $\mathbf{799{\pm}58}$ & $\mathbf{767{\pm}44}$ \\

\cmidrule(lr){2-6}

& \multirow{2}{*}{$\bar{v}$ [m/s]}
& A & $0.64{\pm}0.02$ & $0.59{\pm}0.02$ & $0.50{\pm}0.02$ \\
&
& M & $0.64{\pm}0.01$ & $0.59{\pm}0.02$ & $\mathbf{0.54{\pm}0.02}$ \\

\cmidrule(lr){2-6}

& \multirow{2}{*}{\shortstack{$T_{\mathrm{term}}$\\{[min]}}}
& A & $21.9{\pm}1.5$ & $23.7{\pm}0.9$ & $26.2{\pm}0.8$ \\
&
& M & $\mathbf{19.7{\pm}0.7}$ & $\mathbf{21.7{\pm}1.2}$ & $\mathbf{22.9{\pm}1.3}$ \\

\bottomrule
\end{tabular}
\end{table}

\subsection{Exploration Performance} \label{sec:exploration_performance}

\begin{figure*}[t]
  \centering
  \subfloat[School]{%
    \label{subfig:coverage_school}%
    \includegraphics[width=0.31\textwidth]{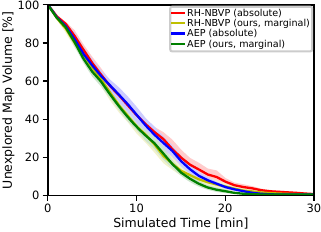}}%
  \hfill
  \subfloat[Large maze]{%
    \label{subfig:coverage_maze}%
    \includegraphics[width=0.31\textwidth]{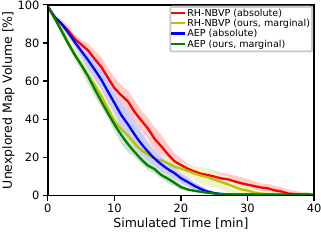}}%
  \hfill
  \subfloat[Multi-story]{%
    \label{subfig:coverage_multistory}%
    \includegraphics[width=0.31\textwidth]{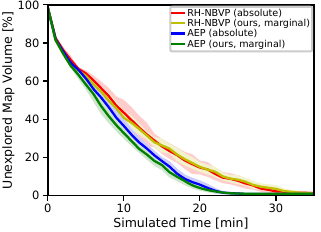}}%
  \caption{Comparison of remaining unexplored volume obtained with RH-NBVP and AEP using absolute and marginal gain in school (a), large maze (b), and multi-story (c) environments.}
  \label{fig:exploration_rate}
\end{figure*}

We compare absolute and marginal gain using RH-NBVP and AEP across the three environments, with the exploration curves in Figure~\ref{fig:exploration_rate}. AEP reaches every coverage milestone earlier with marginal gain in all environments, and the improvement grows with coverage. Between $E_{25}$ and $E_{95}$ it rises from 0.8 to 1.8 min in the school, from 1.3 to 2.3 min in the large maze, and from 0.2 to 1.1 min in the multi-story environment. This is not explained by faster motion, since the average velocities are similar for both formulations. Marginal gain completes exploration with a shorter path in two of the three environments, from 895 to 799 m in the school and 520 to 505 m in the multi-story environment. The large maze keeps a similar traveled distance, at 679 against 684 m, while $E_{95}$ decreases by 2.3 min. AEP also terminates earlier in every environment, by 2.0 min in the school, and 0.5 min in the other environments. The improvement in $E_{95}$ exceeds the improvement in termination time in the large maze, indicating that most of it occurs before high coverage is reached. Afterward, only small and scattered regions remain, resulting in similar termination times.

\begin{figure}[t]
  \centering
  \subfloat[Absolute, $58.8\%$]{%
    \label{subfig:maze_nbvp_absolute}%
    \includegraphics[width=0.43\linewidth]{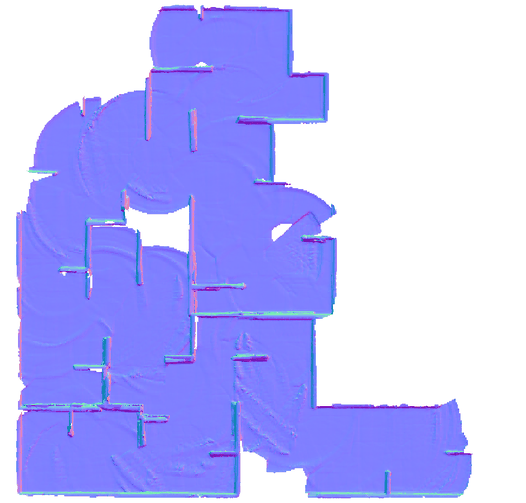}}%
  \hfill
  \subfloat[Marginal, $75.0\%$]{%
    \label{subfig:maze_nbvp_marginal}%
    \includegraphics[width=0.43\linewidth]{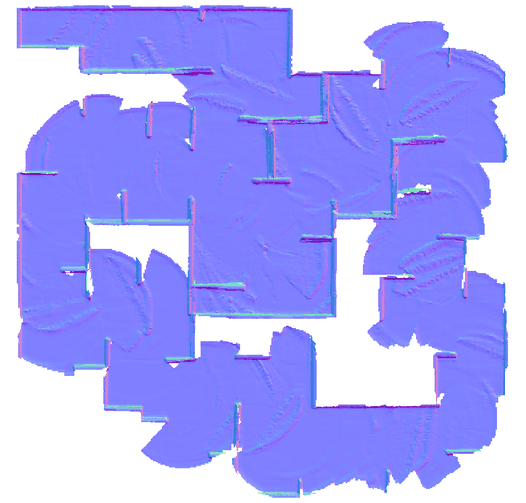}}%
  \caption{Large-maze reconstruction after $15$ min with RH-NBVP using
  absolute (a) and marginal gain (b).}
  \label{fig:maze_reconstruction}
\end{figure}

For RH-NBVP, the improvement depends more on the environment, reducing $E_{95}$ by 0.7 min in the school, and 3.8 min in the large maze. The large maze shows the most consistent improvement, growing from 2.4 min at $E_{25}$ to 3.8 min at $E_{95}$. This is also visible in Fig.~\ref{fig:maze_reconstruction}, where marginal gain increases the explored volume after $15$ min from $58.8\%$ to $75.0\%$. In the school the difference increases to 2.1 min at $E_{75}$ before decreasing to 0.7 min at $E_{95}$. The path length and velocity show no consistent advantage for either formulation.

The multi-story environment is the only case where marginal gain does not improve $E_{95}$ with RH-NBVP. Although it reaches $E_{25}$ and $E_{50}$ earlier, absolute gain becomes 0.4 min faster at $E_{75}$ and $E_{95}$. The tighter layout and the required $d_{\mathrm{collision}}$ make it harder for the sampling tree to reach informative regions, so additional nodes do not necessarily improve access to unexplored areas. The gain formulation can change the rank of reachable viewpoints, but it cannot compensate for limitations in the sampling process. The same effect is visible with AEP, where the termination time differs by only 0.5 min. Final coverage remains similar for both formulations in all environments, indicating that marginal gain improves exploration efficiency but not the final mapped volume.

\subsection{Effect of Execution Horizon} \label{sec:execution_horizon}

We evaluate RH-NBVP in the large maze with execution horizons $h=1$, $3$, and $5$, where $h$ is the number of consecutive nodes executed before replanning.  The large maze is used because its size requires many planning decisions throughout the mission, making the effect of committing to longer branches more evident. Table~\ref{tab:execution_horizon} reports the time $E_{95}$, path length $L_{95}$ and average velocity $\bar{v}_{95}$ to reach $95\%$ coverage, together with the total planning time $T_{\mathrm{plan}}$.

\begin{table}[t]
\centering
\renewcommand{\arraystretch}{0.90}
\caption{Effect of the execution horizon in the large maze.
A/M denote absolute and marginal gain.}
\label{tab:execution_horizon}
\footnotesize
\begin{tabular}{@{}ccrrrr@{}}
\toprule
$\boldsymbol{h}$ &
\textbf{Gain} &
$\boldsymbol{E_{95}}$ \textbf{[min]} &
$\boldsymbol{L_{95}}$ \textbf{[m]} &
$\boldsymbol{\bar{v}_{95}}$ \textbf{[m/s]} &
$\boldsymbol{T_{\mathrm{plan}}}$ \textbf{[s]} \\
\midrule

\multirow{2}{*}{1}
& A & $30.5{\pm}3.7$ & $635{\pm}79$  & $0.35{\pm}0.01$ & $\mathbf{156{\pm}2}$ \\
& M & $\mathbf{26.7{\pm}2.6}$ & $\mathbf{565{\pm}60}$ &
$0.35{\pm}0.01$ & $353{\pm}6$ \\

\midrule

\multirow{2}{*}{3}
& A & $28.9{\pm}2.8$ & $696{\pm}77$  & $0.40{\pm}0.01$ & $\mathbf{69{\pm}4}$ \\
& M & $\mathbf{26.0{\pm}1.5}$ & $\mathbf{635{\pm}39}$ &
$\mathbf{0.41{\pm}0.01}$ & $151{\pm}6$ \\

\midrule

\multirow{2}{*}{5}
& A & $30.1{\pm}5.7$ & $743{\pm}152$ & $0.41{\pm}0.01$ & $\mathbf{55{\pm}4}$ \\
& M & $\mathbf{26.5{\pm}1.8}$ & $\mathbf{659{\pm}46}$ &
$\mathbf{0.42{\pm}0.01}$ & $105{\pm}9$ \\

\bottomrule
\end{tabular}
\end{table}

The results show that $E_{95}$ remains similar as the horizon increases with marginal gain. Longer horizons require fewer replanning iterations, so $T_{\mathrm{plan}}$ decreases from \SI{353}{\second} at $h=1$ to \SI{105}{\second} at $h=5$, while $\bar{v}_{95}$ increases from \SI{0.35}{\meter\per\second} to \SI{0.42}{\meter\per\second}. However, path length increases with the execution horizon. With $h=1$, the planner updates its decision after every executed node using the latest map, resulting in more efficient paths. For longer horizons, the UAV must execute more of the selected branch before changing direction. As a result, $L_{95}$ with marginal gain grows from \SI{565}{\meter} to \SI{635}{\meter} and \SI{659}{\meter}, and absolute gain follows the same trend, from \SI{635}{\meter} to \SI{696}{\meter} and \SI{743}{\meter}. Therefore, longer horizons trade path efficiency for reduced replanning time and higher velocities, keeping $E_{95}$ similar.

More importantly, the benefit of marginal gain persists at longer horizons. At $h=3$ it reaches $95\%$ coverage with \SI{635}{\meter}, the same mean path length as absolute gain at $h=1$, needing just \SI{24}{\meter} more at $h=5$. As such, marginal gain at $h=3$ and $h=5$ achieves similar path efficiency to absolute gain replanning after every node, also reducing $E_{95}$ from \SI{30.5}{\minute} to \SI{26.0}{\minute} and \SI{26.5}{\minute}. At $h=3$, this is achieved with comparable planning times, \SI{151}{\second} versus \SI{156}{\second}.

\section{REAL-WORLD EXPERIMENTAL EVALUATION} \label{sec:real_world}

To validate the proposed method in real-world conditions, we deployed AEP with absolute and marginal gain on a DJI F550 UAV equipped with an Intel RealSense D455 and a Pixhawk flight controller running ArduPilot, utilizing Voxblox for real-time mapping. State estimation is obtained from RTK-GPS and IMU measurements. Mapping, planning, and gain evaluation are run in real time onboard an NVIDIA Jetson Orin NX with 16~GB of unified memory. The experiments were performed in a Patio environment ($10\times13\times6$ m), an outdoor area containing a lamp structure and two flower beds separated by a staircase. The main parameters were set to $v_{xy_{\max}}=v_{z_{\max}}=0.5$~m/s, $g_{\mathrm{zero}}=1$~m$^3$, $N_{\max}=200$, and $N_{\mathrm{termination}}=400$ across conditions.

The results show the same trend observed in simulation. Marginal gain reaches $50\%$, $75\%$, and $95\%$ coverage in $0.49{\pm}0.05$, $0.92{\pm}0.10$, and $2.17{\pm}0.44$~min, compared with $0.54{\pm}0.08$, $1.01{\pm}0.19$, and $3.10{\pm}1.05$~min with absolute gain. The difference is largest at $95\%$ coverage, where marginal gain reduces the exploration time by $30\%$. Termination time is also reduced from $4.58{\pm}0.46$ to $4.04{\pm}0.34$~min, with path length decreasing from $108.5{\pm}11.2$ to $101.4{\pm}7.0$~m and average velocity remaining similar at $0.34{\pm}0.05$ and $0.36{\pm}0.01$~m/s. This shows that the improvement does not rely on faster UAV motion. Figure~\ref{fig:patio_reconstruction} shows the corresponding RTAB-Map \cite{Labb_2018} reconstructions for visualization.

\section{CONCLUSION} \label{sec:conclusion}

We presented a GPU-accelerated method for computing path-dependent marginal information gain using depth buffers and depth-ordered tree evaluation. Compared with the exact marginal gain that uses voxel hash maps, the proposed implementation stays within $5$-$10\%$ of the exact value while achieving speed-ups of up to $118\times$. Experiments with RH-NBVP and AEP showed that marginal gain reached $95\%$ coverage earlier in five of the six planner-environment combinations. The benefit also increased with larger trees and remained effective for longer execution horizons. Finally, real-world experiments reproduced the same behavior onboard a UAV, where marginal gain reduced the time to $95\%$ coverage by $30\%$ and also achieved earlier termination and a shorter path. For future work, we propose extending the method to other exploration planners.


\bibliographystyle{IEEEtran}
\bibliography{references}

\end{document}